\documentclass[conference]{IEEEtran}

\usepackage{cite}

\ifCLASSINFOpdf
\else
\fi

\usepackage{url}

\usepackage{graphicx}
\usepackage{booktabs} % for booktabs table
\usepackage{multirow}
\usepackage{enumitem} 
\usepackage[dvipsnames]{xcolor}
\usepackage{csquotes}

\begin{document}

\title{Utilizing a Transparency-driven Environment toward Trusted Automatic Genre Classification: \\ A Case Study in Journalism History}
\author{\IEEEauthorblockN{Aysenur Bilgin, Laura Hollink, \\Jacco van Ossenbruggen}
\IEEEauthorblockA{CWI, Amsterdam\\
\{aysenur.bilgin, l.hollink, \\ jacco.van.ossenbruggen\}@cwi.nl}
\and
\IEEEauthorblockN{Erik Tjong Kim Sang}
\IEEEauthorblockA{Netherlands eScience Center\\
e.tjongkimsang@esciencecenter.nl}
\and
\IEEEauthorblockN{Kim Smeenk, Frank Harbers, \\Marcel Broersma}
\IEEEauthorblockA{University of Groningen\\
\{k.s.p.smeenk, f.harbers, \\ m.j.broersma\}@rug.nl}
}
\IEEEoverridecommandlockouts
\IEEEpubid{\makebox[\columnwidth]{978-1-5386-5541-2/18/\$31.00~\copyright2018 IEEE \hfill} \hspace{\columnsep}\makebox[\columnwidth]{ }}
% make the title area
\maketitle
\IEEEpubidadjcol
\begin{abstract}
With the growing abundance of unlabeled data in real-world tasks, researchers have to rely on the predictions given by black-boxed computational models. However, it is an often neglected fact that these models may be scoring high on accuracy for the wrong reasons. In this paper, we present a practical impact analysis of enabling model transparency by various presentation forms. For this purpose, we developed an environment that empowers non-computer scientists to become practicing data scientists in their own research field. We demonstrate the gradually increasing understanding of journalism historians through a real-world use case study on automatic genre classification of newspaper articles. This study is a first step towards trusted usage of machine learning pipelines in a responsible way.

\end{abstract}

\IEEEpeerreviewmaketitle

\section{Introduction}

Genre is an important attribute for studying the development of newspapers over time \cite{broersma2007} \cite{broersma2008discursive} \cite{harbers2014}. However, in contrast to topic, information about genre cannot be found using key word search, nor is fine-grained genre information readily available in historical newspaper collections. Therefore, it needs to be added. However, due to the growing size of the data sets in journalism history research \cite{broersma2011}, manual categorization has become infeasible. In 2017, Harbers and Lonij \cite{harbers2017} showed that automatic prediction of genre labels for Dutch newspaper articles is possible with a reasonable labeling accuracy. 

From a machine learning perspective, the aim is to improve the accuracy of the prediction model. However, as George Box wrote: \enquote{All models are wrong} \cite{box1976}. Prediction accuracy, albeit carefully evaluated, may not be able to provide a complete assessment of the performance of a classification model, especially if it deals with multiple classes whose distribution over the data sets is unbalanced. On top of this, from a journalism history perspective, knowing what type of errors the classification model makes is crucial for being able to assess whether machine learning model predictions can be trusted. To unite the two perspectives in a single environment, we worked out various types of comprehensible presentations to analyze the decisions of machine learning models. By thus increasing the transparency of the models, we support the journalism historians in deciding which model's predictions are best for enhancing their research. 

As a case study for the transparency-driven environment, we present a real-world scenario using hypotheses from journalism history that can be drawn from applying an automatic method for genre classification of newspaper articles on large-scale unlabeled data. The objective is not only performing well on performance metrics but also being able to explain the predictions of the machine learning pipelines to the journalism historians. 

This paper is organized as follows: In Section \ref{sec-relatedwork}, we provide background on transparency in machine learning. Section \ref{sec-environment} presents the design of an environment for transparent machine learning pipelines. The data sets together with the real-world application on journalism history are introduced in Section \ref{sec-usecase}. Next, Section \ref{sec-discussion} contains a discussion on the challenges and the lessons learned. Finally, we note some concluding remarks and lay out the future work in Section \ref{sec-conclusion}.

\section{Background and Related work}
\label{sec-relatedwork}
The emergence of large open newspaper repositories such as the Dutch Delpher\footnote{\url{https://delpher.nl/}} \cite{smits2015} makes millions of newspaper articles available for research. While the abundance of data is fueling data-driven approaches, essentially machine learning, it is also bringing along concerns. Irresponsible data usage and the concealing nature of automation in decision making processes are among these concerns \cite{duarte2018}\cite{vanderAalst2017}. Initiatives such as the RDS\footnote{\url{http://www.responsibledatascience.org/}} and Explainable AI (XAI) \cite{gunning2017explainable} aim to tackle these concerns by increasing awareness on transparency and encouraging the development of new methods to help humans understand the models and appropriately trust them. Furthermore, artificial intelligence, and machine learning in particular, has begun to have progressively more impact in people's lives. In response, regulations concerning transparency have made their entry to European Union's General Data Protection Regulation (GDPR).\footnote{\url{https://gdpr-info.eu/art-22-gdpr/}}

The surge of interest in transparency is tightly coupled with the notion of interpretability. Both concepts have different uses and non-overlapping motivations throughout the literature \cite{lipton2016mythos}. According to Lipton \cite{lipton2016mythos}, transparency is a property of an interpretable model and is examined on three levels: 1) entire model (simulatability), 2) individual components (decomposability), and 3) training algorithm (algorithmic transparency). For the scope of this paper, we consider decomposability to achieve transparency in our analysis as the parameters of the model play an important tangible role in uncovering whether the model's intentions resonate with the theory used for the extrinsic task. Regarding interpretability, we consider the definition provided by Doshi-Velez and Kim \cite{doshi2017towards} and refer to the term as the ability to explain or to present in understandable terms to a human how the machine learning system decides. For the use case study in which we collaborate with journalism historians, it is important to expose for which examples the machine learning decisions are right. Among the approaches for such post-hoc interpretability outlined in Lipton \cite{lipton2016mythos}, we make use of visualizations (e.g. tabular views) and local explanations (e.g. LIME \cite{ribeiro2016should}). There are various open source systems/packages that aim to create, understand, debug, optimize and monitor machine learning pipelines such as TensorBoard\footnote{\url{https://www.tensorflow.org/guide/summaries_and_tensorboard}}, Spark ML\footnote{\url{https://spark.apache.org/docs/1.2.2/ml-guide.html}} and Skater\footnote{This project is in beta phase and at the time of writing still heavily under development} \cite{pramit_choudhary_2018_1198885}. However, the common objective of these architectures is to enrich the understanding of either the developer or the data scientist whereas in our work, the understanding of the journalism historian lies at the core.

\section{Design of an Environment for Transparent Machine Learning Pipelines}
\label{sec-environment}
In this section, we present the components of an environment that advocates trust and promotes transparency. 
We have two major objectives for functionality, one is to support the journalism historian by providing detailed insights on the utilization of machine learning pipelines and the other is to facilitate the comparison of performance metrics for the pipelines created using the environment. The machine learning pipelines we consider in this study begin with the data set selection and pre-processing configuration. Presumably, there have been various decisions made while compiling the data sets we use, however, for the scope of this paper, we exclude the details and methods employed for data collection and annotation. Since the data sets relate to the use case study, we provide further details for them in Section \ref{subsec-dataset}. Furthermore, we note that the environment has the flexibility to integrate a variety of algorithms, text processing tools and techniques, which may not be mentioned in this paper due to time and space limitations. 

\begin{figure}[t]
\includegraphics[scale=0.13]{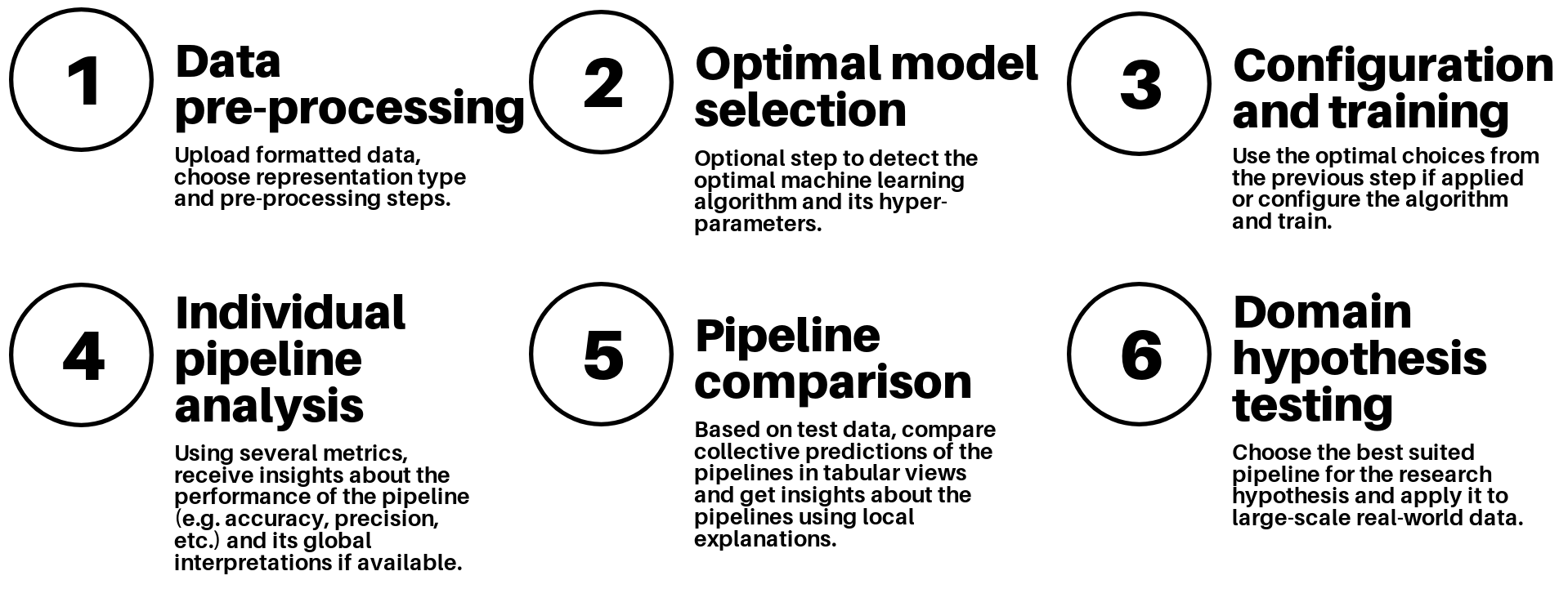}
\caption{The workflow of turning data into robust claims in a transparent way. The combination of Step 1 and Step 3 is referred to as ML pipeline.}
\label{fig-workflow}
\end{figure}

The workflow of creating and using a machine learning (ML) pipeline in the environment contains six steps as illustrated in Fig. \ref{fig-workflow} and as described in the following subsections. 

\subsection{Step 1: Data pre-processing}
We consider three representation types for the documents to be used in the ML pipelines. The representations also include pre-processing of the document as detailed below:
\label{subsec-datarepr}
\begin{enumerate}[label=(\roman*)]
	\item  Term Frequency Inverse Document Frequency (TF-IDF): We convert the raw documents to a matrix of TF-IDF scores with the function TfidfVectorizer\footnote{\url{http://scikit-learn.org/stable/modules/generated/sklearn.feature_extraction.text.TfidfVectorizer.html}} from the Python package scikit-learn \cite{scikit-learn}. TF-IDF scores incorporate the frequency of a term in a document (TF) as well as the inverted document frequency (IDF), which is equal to one divided by the number of documents that contain the term. We use four non-default settings for the vectorizer function:

\textit{sublinear\_tf} is set to \textit{True} to use a logarithmic form for frequency, \textit{min\_df} is set to \textit{5} as the minimum number of documents a word must be present in to be kept in the vocabulary, \textit{norm} is set to \textit{'l2'} to ensure all our feature vectors have a euclidean norm of 1, and \textit{ngram\_range} is set to \textit{(1, 2)} to include both unigrams and bigrams.
    
    \item TF-IDF with stop-word removal (TF-IDF (SWR)): As a pre-processing technique, we apply stop-word removal before extracting TF-IDF features from the documents. The stop-word list can be customized according to the use case as further detailed in Section \ref{subsec-datapreproc}.
    
    \item \textless{}NLP suite\textgreater{} with scaling (\textless{}NLP suite\textgreater{} (SCL)): By using natural language processing (NLP) tools, we can parse the documents and extract sentences, named entities, part-of-speech tags, etc. that lead to manual curation of various numerical features as further detailed in Section \ref{subsec-datapreproc}. It is possible to customize the list of these features as well as selecting which ones to be included in further analysis. When dealing with numerical features, some of the machine learning algorithms may require scaling of the data. For this purpose, we employ RobustScaler\footnote{\url{http://scikit-learn.org/stable/modules/generated/sklearn.preprocessing. RobustScaler.html}} from scikit-learn \cite{scikit-learn}, which scales features using methods that are robust to outliers. According to the naming convention we use, the name of the NLP suite will be indicated in the brackets. As an example, the representation type will be referred to as FROG (SCL) when we employ Frog\footnote{\url{http://languagemachines.github.io/frog/}} as the NLP suite.

\end{enumerate}

\subsection{Step 2: Optimal model selection}
This can be considered to be an optional step. Herein, we consider different supervised learning methods to build models of the available data including but not limited to Support Vector Machines, Naive Bayes, and Random Forests. 

The optimization is done using scikit-learn's GridSearchCV \footnote{\url{http://scikit-learn.org/stable/modules/generated/sklearn.model_selection.GridSearchCV.html}} with the following specified parameters values per machine learning algorithm: 
\begin{itemize}
\item Support Vector Classifier (SVC) \cite{vapnik1995}: We opted to include the kernel, C and gamma hyper-parameters in the grid search with the values of {\lq}linear{\rq} and {\lq}rbf{\rq} for the kernel, [1, 2, 3, 5, 10] for the penalty parameter C, and [0.1, 0.01, 0.001] for gamma, which is the kernel coefficient for {\lq}rbf{\rq}. 
\item Multinomial Naive Bayes (NB) \cite{minsky1961}: We decided to include the values [0, 0.01, 0.02, 0.05, 0.1, 0.2, 0.5, 1.0] for the smoothing hyper-parameter alpha.
\item Random Forest Classifier (RF) \cite{ho1995}: We chose the Gini impurity (i.e {\lq}gini{\rq}) for the quality of the split. We varied the number of estimators and the maximum features to be [50, 100, 1000] and [1, 2, 3, 4, 5, 6], consecutively. 
\end{itemize}

Other machine learning algorithms, such as deep learning (multilayer perceptron \cite{rumelhart1986} in Keras\footnote{https://keras.io}), are being considered for addition to the research environment. We used four different scorers\footnote{The details can be found on \url{http://scikit-learn.org/stable/modules/model_evaluation.html}} (i.e. accuracy, precision (micro), recall (micro) and F1 score (micro)), which are suitable for the multi-class classification problems.

\subsection{Step 3: Configuration and training}
This step is composed of three actions: the first one is selecting features when the representation type of the documents allows for it (i.e. see (\textless{}NLP suite\textgreater{} (SCL)) representation in Section \ref{subsec-datarepr}), the second action is configuring the hyper-parameters of the learning algorithm (that may be suggested in Step 2 or may be carried out as desired by the user), and the third one is training. 

The combination of Step 1 and Step 3 is referred to as a machine learning (ML) pipeline.
 
\subsection{Step 4: Individual pipeline analysis}
We make use of visual graphs that can dynamically accommodate multiple classes. One of the most important visualization shown under the pipeline performance is the confusion matrix. We make use of heat map illustration (see Fig. \ref{fig-cm}) to demonstrate the performance through both 10-fold cross validation and testing on unseen data. The purpose is to carry out model checking using 10-fold cross validation to expose the potential classification performance. In order to ensure transparency on the pipeline's performance, we train the models on 90\% of the data selected for the pipeline and we produce another confusion matrix using the remaining 10\% of the data, which is unseen.

Furthermore, we display the numerical values of well-established metrics such as accuracy, (weighted, micro and macro) precision, recall and F1 score resulting from both 10-fold cross validation and testing of the model on unseen data. This step also allows the investigation of global interpretability, which implies knowing about the patterns present in general \cite{doshi2017towards}, exclusively for linear algorithms. For this purpose, we show feature importance ranking plots (see Fig. \ref{fig-nieuws-vs-rep} and Fig. \ref{fig-rep-vs-col}). For some algorithms (i.e. SVC), the number of these plots depend on the number of classes in the classification task.    

\subsection{Step 5: Pipeline comparison}
In order to be able to critically evaluate the machine learning pipelines and determine the most suited pipeline for the extrinsic task, we designed three tabular views: 1) explanation view, 2) set-based agreement view, and 3) article-based agreement view. The tabular view is populated using the pipelines in the environment applied to the test sets, which are collocated from the database with duplicates removed. Hence, each document will appear once and will have a unique identifier that can be used to trace in different views. We detail the purpose of these views below:

\subsubsection{Explanation View}
In this view, the main purpose is to communicate the local interpretability, which implies knowing
the reasons for a specific labeling decision \cite{doshi2017towards}, retrieved using LIME \cite{ribeiro2016should}. The view depicts the unique document identifier, the pipeline used for the explanation, the prediction for the classification task using the pipeline and textual or tabular explanation depending on the representation type as part of the pipeline. Further analysis on the local explanations is depicted using visualizations provided by LIME \cite{ribeiro2016should} (see Fig. \ref{fig-lime239}). Table \ref{table-explanation-view} represents the structured information shown in this view.

\subsubsection{Set-based Agreement View}
For a binary classification task, it is convenient to use Venn diagrams when comparing up to three pipelines. However, for multi-class classification tasks, these diagrams would not be as useful. Therefore, we developed a set-based agreement view that can accommodate multiple classes and multiple pipelines to be compared. The view depicts mutually agreeing pipelines on the prediction and the magnitude of the agreement. For data sets that have the ground truth, the documents can be matched against and color-coded with respect to the agreement between the prediction and the ground truth. The \textcolor{OliveGreen}{\textless{}green\textgreater{}} documents indicate a correct classification, whereas the \textcolor{BrickRed}{\{red\}} documents are misclassified according to the ground truth labels. The documents that belong to unlabeled data sets are colored \textcolor{Blue}{[blue]}. Table \ref{table-set-based-view} represents the structured information shown in this view.

\subsubsection{Document-based Agreement View}
In the set-based agreement view, the mutually agreeing pipelines serve as index. In this view, however, the documents themselves serve as index. We display mutually agreeing pipelines for the specific document together with the prevailing prediction given by the pipelines and the ground truth if the data set allows for it. Table \ref{table-article-based-view} represents the structured information shown in this view.

\subsection{Step 6: Domain hypothesis testing}
For the domain researchers, the eventual goal for using ML pipelines is to test their hypotheses. In this step, the environment enables the domain researchers to test their hypotheses using plots for a visual analysis. An example of a hypothesis testing plot based on output label frequencies, is given in Fig. \ref{fig-recomputed}.

\begin{table}[t]
\centering
\caption{Explanation View Representation}
\label{table-explanation-view}
\begin{tabular}{@{}ccccc@{}}
\toprule
& & & Textual & Feature \\
Document \#                         & Pipeline                            & Prediction                       & Explanation & Explanation \\ \midrule
\textless{}A\textgreater{} & \textless{}P\textgreater{} & \textless{}Class C\textgreater{} & Click here & N/A                 \\
\textless{}B\textgreater{} & \textless{}Q\textgreater{} & \textless{}Class B\textgreater{} & N/A                 & Click here \\ \bottomrule
\end{tabular}
\end{table}

\begin{table}[t]
\centering
\caption{Set-based Agreement View Representation where label accuracy is represented in the documents column as: \textcolor{OliveGreen}{\textless{}correct\textgreater{}}, \textcolor{BrickRed}{\{wrong\}} and \textcolor{Blue}{[unknown]}.}
\label{table-set-based-view}
\begin{tabular}{@{}cccl@{}}
\toprule
Mutually &&Number&\\
Agreeing &&of&\\
Pipelines & Prediction & documents & \multicolumn{1}{c}{Documents}     \\ \midrule
\begin{tabular}[c]{@{}c@{}}\textless{}B\textgreater\\ \textless{}D\textgreater\\ \textless{}G\textgreater\\ \textless{}H\textgreater{}\end{tabular} & \textless{}Class F\textgreater{} & \textless{}N\textgreater{} & \begin{tabular}[c]{@{}l@{}}
\textcolor{OliveGreen}{\textless{}A\textgreater{}} \textcolor{OliveGreen}{\textless{}K\textgreater{}} \textcolor{OliveGreen}{\textless{}L\textgreater} \textcolor{OliveGreen}{\textless{}M\textgreater}\\ \textcolor{OliveGreen}{\textless{}N\textgreater} \textcolor{BrickRed}{\{P\}} \textcolor{OliveGreen}{\textless{}T\textgreater} \textcolor{BrickRed}{\{U\}}\\ \textcolor{Blue}{[W]} \textcolor{Blue}{[Y]}\end{tabular} \\ \bottomrule
\end{tabular}
\end{table}

\begin{table}[t]
\centering
\caption{Document-based Agreement View Representation}
\label{table-article-based-view}
\begin{tabular}{@{}ccccc@{}}
\toprule
&& Mutually &&\\
&& Agreeing && True\\
Document \#  & Document text & Pipelines & Prediction & Class \\ \midrule
\textless{}A\textgreater{} & \textless{}Text\textgreater{} & \begin{tabular}[c]{@{}c@{}}\textless{}B\textgreater\\ \textless{}D\textgreater\\ \textless{}G\textgreater\\ \textless{}H\textgreater{}\end{tabular} & \textless{}Class F\textgreater{} & \textless{}Class F\textgreater{}\\
\bottomrule
\end{tabular}
\end{table}

Our machine learning models will produce incorrect output label distributions. Fortunately, we can use the performance of the gold standard data to find out what kind of errors the model makes. This assessment can then be used for re-estimating the label counts on unseen data. The predicted counts will include false positives, which can be removed by multiplying them with the precision score for the relevant label. The predicted counts will miss false negatives, which can be compensated for through dividing them by the recall score for a label. Accordingly, we can re-estimate the output label counts using the following formula:

\begin{equation}
New\_count_{label} = Old\_count_{label} * \frac{Precision_{label}}{Recall_{label}}
\end{equation}

Note that the precision and recall scores used for re-estimation are only dependent on the output label and not on the input features. 

\section{Use-case Study: Automatic Genre Classification in Journalism History}
\label{sec-usecase}
This section elaborates on the use-case study to demonstrate the environment presented in Section \ref{sec-environment}. To begin with, we provide background on the automatic genre classification task in journalism history. Next, we present the motivation and hypotheses of this use-case study. Then, we provide the details of the data sets available in the environment and continue with the analysis performed by applying the last two steps of the workflow (i.e. Step 5 and Step 6 of Section \ref{sec-environment}).

\subsection{Automatic Genre Classification in Journalism History}

\begin{table*}
\centering
\caption{Results for application of Step 2 of the workflow: Optimized algorithms and hyper-parameters for the data sets using GridSearchCV (SVC: Support Vector Classifier, RF: Random Forest)}
\label{dataset_algorithms_overview}
\begin{tabular}{@{}ccccccc@{}}
\toprule
            &\multicolumn{2}{c}{BGS}    & \multicolumn{2}{c}{GS}   & \multicolumn{2}{c}{CGS}  \\ 
            & ALG & HP & ALG & HP & ALG & HP \\ \midrule
\multirow{2}{*}{TF-IDF}        & SVC & kernel: rbf  & SVC & kernel:linear  & SVC & kernel:linear \\
              &  &  C:10  &  & C:3  &  & C:3 \\
\multirow{2}{*}{TF-IDF (SWR)} & SVC & kernel:linear & SVC & kernel:linear & SVC & kernel:linear \\
              &  & C:2 &  & C:3 &  & C:3 \\
\multirow{2}{*}{FROG (SCL)}  & RF & n\_estimators:50 & RF & n\_estimators:1000 & RF & n\_estimators:1000 \\
               &  &  max\_features:5 &  &  max\_features:5 &  & max\_features:2  \\ \bottomrule
\end{tabular}
\end{table*}

The genre distribution over time gives journalism historians insight into the development of journalism. De Melo \& De Assis \cite{marques2016generos} have proposed a classification model of journalistic genres consisting of five genres: informative, opinionative, interpretative, diversional and utility. These genres are subdivided into formats, such as the editorial. Their model is based on the communicative purpose of the text. De Haan-Vis \& Spooren \cite{de2016informalization} defined "journalistic prose" as a genre that consists of multiple subgenres, e.g. news reports and interviews. Their genre classification model is based on the (professional) context of the text and the mode of the text. Broersma, Harbers \& Den Herder have differentiated between 18 genres, which are distinguished based on their textual form \cite{harbers2014}. They have defined the genre labels by looking at how genres have been talked about historically by journalists, how they are defined in journalistic handbooks and self-classified in newspapers. De Melo \& De Assis' formats and De Haan-Vis \& Spooren's subgenres largely overlap with the genres Broersma, Harbers \& Den Herder have distinguished. In this paper, we follow the work of the latter as it enables the domain scientists to get insight into how genres have developed throughout the history. Particularly, we work with the following genres: News (in Dutch: {\it Nieuwsbericht}), Background ({\it Achtergrond}), Report ({\it Verslag}),  Interview ({\it Interview}),  Feature ({\it Reportage}),  Op-ed ({\it Opiniestuk}),  Review ({\it Recensie}), and Column ({\it Column}). 

In 2017, Harbers and Lonij \cite{harbers2017} showed that automatic prediction of genre labels for Dutch newspaper articles is possible with a reasonable labeling accuracy. They used the machine learning method Support Vector Machines \cite{boser1992} to build classification models from manually labeled newspaper articles and achieved 65\% accuracy. In this study, we refer to the results in Harbers and Lonij \cite{harbers2017} as a baseline, and to showcase the potential value of a transparent environment, we focus on the scientific understanding of the domain scientist.

\subsection{Motivation and Hypotheses}
\label{sec-hypotheses}
From the end of the 19th century onward, journalism moved from partisan, opinion-oriented journalism to an independent, fact-centered journalism practice \cite{harbers2014}. Based on our primary interest in the development of opinionated and fact-centered genres, we look at two hypotheses: 
\begin{enumerate}
\item The relative amount of \textit{Reports} increased in 1985 compared to 1965.
\item The relative amount of \textit{Features} increased in 1985 compared to 1965.
\end{enumerate}

Like Wevers et al. \cite{wevers2018} do for distributions of illustrations, we will create graphs of distributions of the genre labels over time, both for the gold-standard labels and for the predictions of the machine learning pipelines for unlabeled data. We will test if the graphs satisfy these two hypotheses.

\subsection{Data sets}
\label{subsec-dataset}
The raw data sets that were made available for this study are detailed below:
\begin{itemize}
\item BGS (Balanced Gold Standard)
\item GS (Gold Standard)
\item CGS (Combined Gold Standard)
\item UD (Unlabeled Data)
\end{itemize}

We assume that the articles in the data sets have been selected randomly and that data selection bias was avoided (but see also Traub et al. \cite{traub2016}). BGS is a balanced data set, which contains 60 articles of each of the eight genres of interest (\textit{Background, Column, Interview, News, Op-ed, Report, Review} and \textit{Feature}). In total, 480 articles were taken from nine different Dutch national newspapers (\textit{Gereformeerd Gezinsblad, Nederlands Dagblad, Algemeen Handelsblad, NRC, Parool, Telegraaf, Volkskrant, Vrije Volk, Waarheid}) and from five years (1955, 1965, 1975, 1985 and 1995). The label distribution over the years is uniform. The manual annotation was performed by a single person.

GS is based on the data set created by Broersma, Harbers \& Den Herder \cite{harbers2014}. The original data set was a collection of metadata built at a time when no digital versions of the newspapers were available. We used only articles from the years 1965 and 1985 from the newspaper NRC. Not all of the carefully collected metadata could be linked to the digital articles. The 1,424 articles that could be linked automatically \cite{harbers2017} are stored in the Gold Standard (GS) set while the others (884 articles) have been stored in the Unlabeled Data (UD) set. The articles in GS  were  labeled  by  various  annotators  but  each  article  was labeled only by a single annotator. Harbers \& Den Herder \cite{harbers2014} measured the intercoder agreement for a small part of the data set and found an agreement of 77\% (Krippendorf's alpha is 0.67). 

CGS is a combination of the two data sets BGS and GS. The order of documents in all of the data sets is randomized. Since BGS, GS, and CGS sets have annotations, we use them both for training and evaluation of the genre classifiers whereas UD is solely used for simulating large-scale real-world data, which would be used for the eventual hypotheses testing.

\subsection{Data pre-processing and feature selection}
\label{subsec-datapreproc}
For document representation type (TF-IDF (SWR)) in which we remove stop-words, we use a modified list of stop-words based on the Dutch stop-words retrieved from the stop-words\footnote{\url{https://pypi.org/project/stop-words/}} Python project. The modified list (of 86 words) consists of the original 101 words minus the 15 personal pronouns {\lq}haar{\rq}, {\lq}hem{\rq}, {\lq}hij{\rq}, {\lq}hun{\rq}, {\lq}ik{\rq}, {\lq}je{\rq}, {\lq}me{\rq}, {\lq}men{\rq}, {\lq}mij{\rq}, {\lq}mijn{\rq}, {\lq}ons{\rq}, {\lq}u{\rq}, {\lq}uw{\rq}, {\lq}ze{\rq} and {\lq}zij{\rq}. Personal pronouns are excluded from the stop-word list because from a journalism history perspective the distinction between several genres is, among others, based on whether there is a focus on the experience or opinion of the writer. The personal pronoun {\lq}ik{\rq} (I) is indicative of this. 

Regarding feature selection, we employ structural cues such as syntactic categories \cite{kessler1997automatic} (e.g. part-of-speech tags) and use manually curated features based on the definitions of the relevant genres in the codebook for manual content analysis by Broersma, Harbers \& Den Herder \cite{harbers2017}. These features require parsing and tagging of the text by using NLP tools. For Dutch language, we use an easy-to-use NLP suite known as Frog \cite{bosch2007efficient}. The manually curated features extracted from a document are numerical as they represent frequency counts such as the number of adjectives, the number of modal verbs, the number of sentences etc. We also use features such as subjectivity and polarity both of which are retrieved using the module pattern.nl.\footnote{\url{https://www.clips.uantwerpen.be/pages/pattern-nl}} The total number of manually curated features that we consider for this paper is 38.

By the end of the feature extraction and pre-processing steps, we have 9 different training data sets which are listed in Table \ref{dataset_algorithms_overview}. For the operation of the pipelines, we divide each data set into training (90\%) and test (10\%) sets using the same random seed so that the test sets for the same data set will contain the same data items.

\begin{figure}[t]
\includegraphics[scale=0.25]{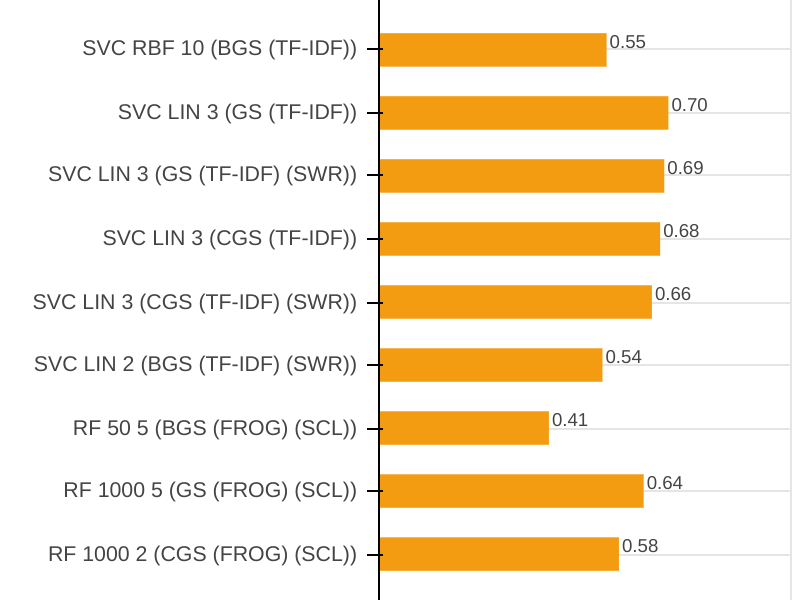}
\caption{Accuracy scores of the ML pipelines available in the environment. The names of the pipelines consist of the algorithm name (RF/SVC) followed by two of its hyper-parameters (50 5/1000 2/1000 5/LIN 2/LIN 3/RBF 10) and then between brackets the training data set (BGS/CGS/GS) and the pre-processing method (FROG/TF-IDF). The use of stop-word removal and scaling is indicated by including SWR or SCL at the end of the pipeline name.}
\label{fig-accuracy}
\end{figure}

\begin{figure*}[t]
\includegraphics[scale=0.18]{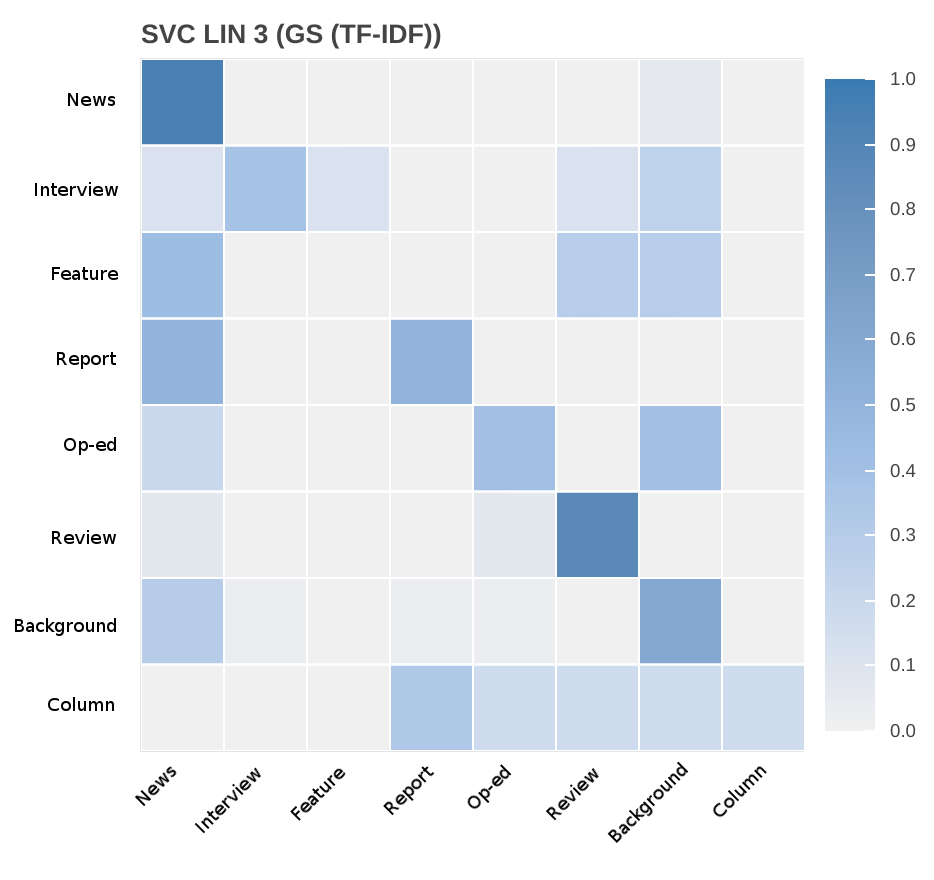}
\includegraphics[scale=0.18]{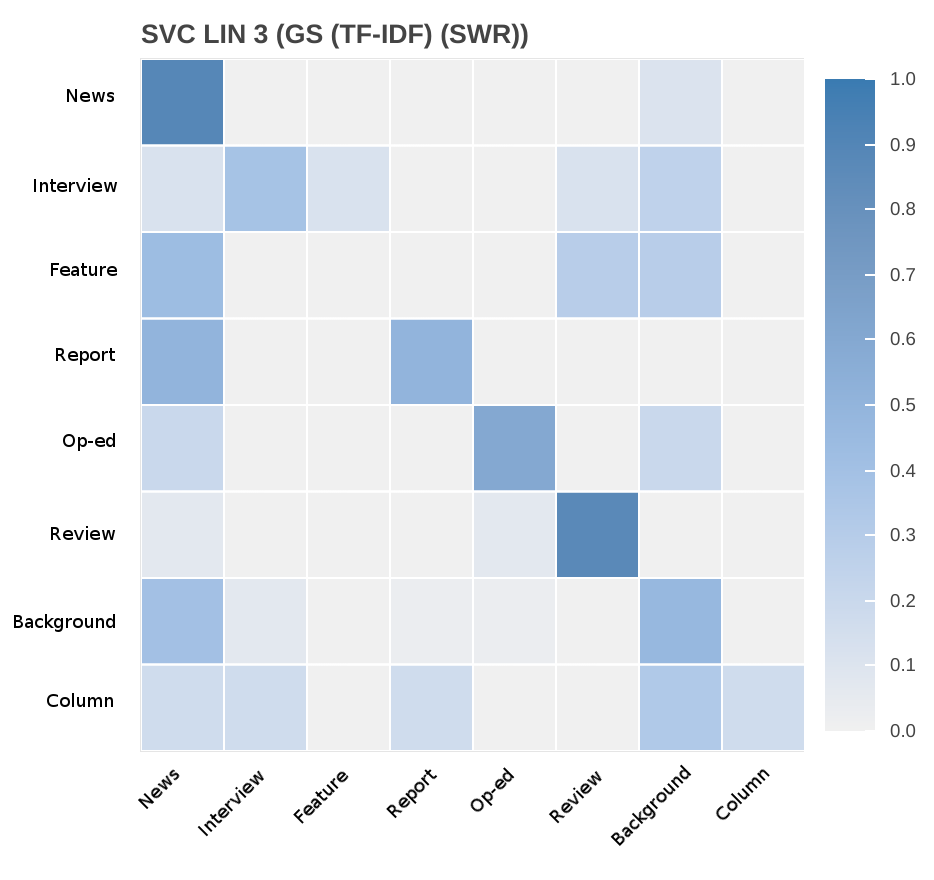}
\includegraphics[scale=0.18]{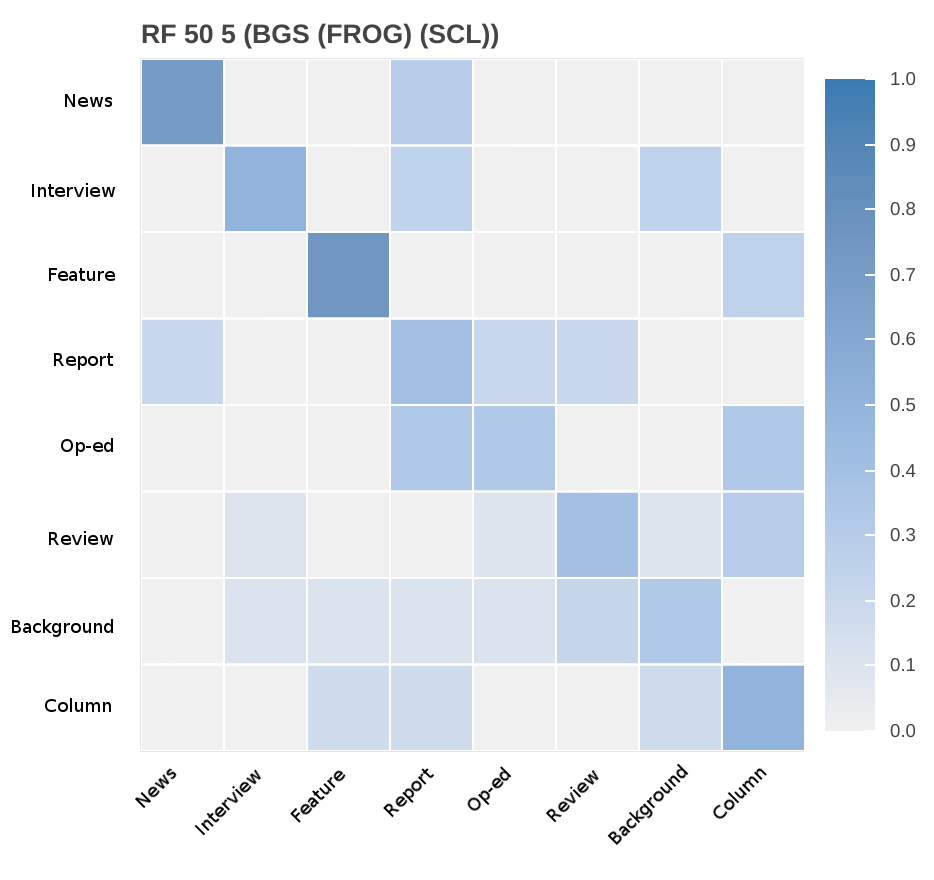}
\caption{Confusion matrices for the selected ML pipelines (populated using only the unseen 10\% of the data set)}
\label{fig-cm}
\end{figure*}

\subsection{Pipeline comparison}
Comparing various pipelines in a systematic way as offered by the transparency-focused environment helps the domain scientist choose the best available one for hypothesis testing. For this study, we chose to investigate 9 different machine learning pipelines that were created using the optimal algorithms and hyper-parameters from Step 2 of the workflow as shown in Table \ref{dataset_algorithms_overview}. 

In the beginning, we observed that the performance measure accuracy\footnote{Computed using accuracy\_score from scikit-learn (see \url{http://scikit-learn.org/stable/modules/generated/sklearn.metrics.accuracy_score.html)}} was the most appealing selection criterion for the domain scientist when evaluating a machine learning pipeline. From a journalism historian point of view, the pipeline named as SVC LIN 3 (GS (TF-IDF)) looks the most promising (see Fig. \ref{fig-accuracy}) followed by the pipeline SVC LIN 3 (GS (TF-IDF) (SWR)), which is trained on the same data set with stop-words removed using the same algorithm and hyper-parameters. 

Due to space limitations, we had to limit our further analysis to include three pipelines. In order to choose the candidate pipelines, we make use of the confusion matrices that give further insights on the individual class-level performance. We observe that although the pipeline RF 50 5 (BGS (FROG) (SCL)) scores the lowest accuracy as shown in Fig. \ref{fig-accuracy}, the confusion matrix is noted by the journalism historian to be favorable, because it has the most equal distribution on the diagonal compared to others. For the rest of the paper, we look at these three ML pipelines whose confusion matrices are illustrated in Fig. \ref{fig-cm}.

Examining the diagonal of the confusion matrices of the two SVC pipelines in Fig. \ref{fig-cm}, we see that the classification performance of \textit{News} and \textit{Review} are relatively better than the rest of the genres. However, the performance for the class \textit{Feature} is unfavorable. We interpret this as these models' failure to classify any of the \textit{Feature} articles correctly in the test set. Since one of our hypotheses is focused on \textit{Feature}, it is crucial to observe good performance in this class when determining the most appropriate pipeline for a robust testing of the hypothesis. We note that RF pipeline in Fig. \ref{fig-cm} can provide a fair comparison in this regard.

\subsubsection{Global interpretations}
\begin{figure}[t]
\includegraphics[scale=0.23]{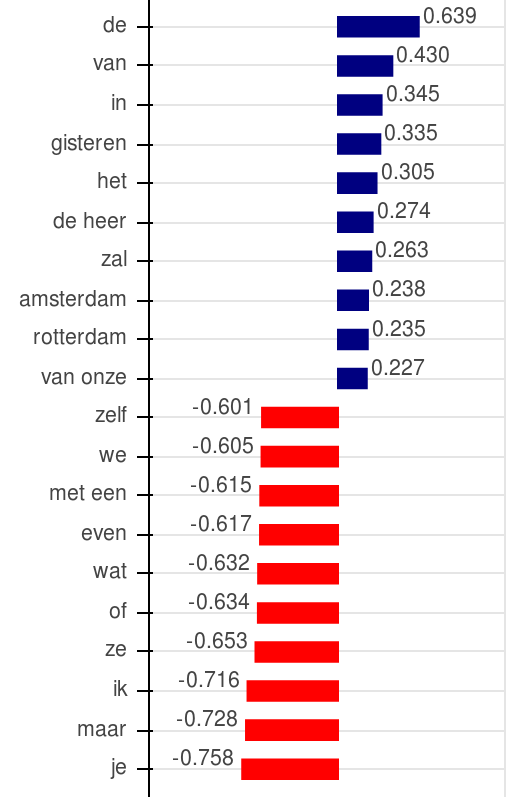}
\includegraphics[scale=0.23]{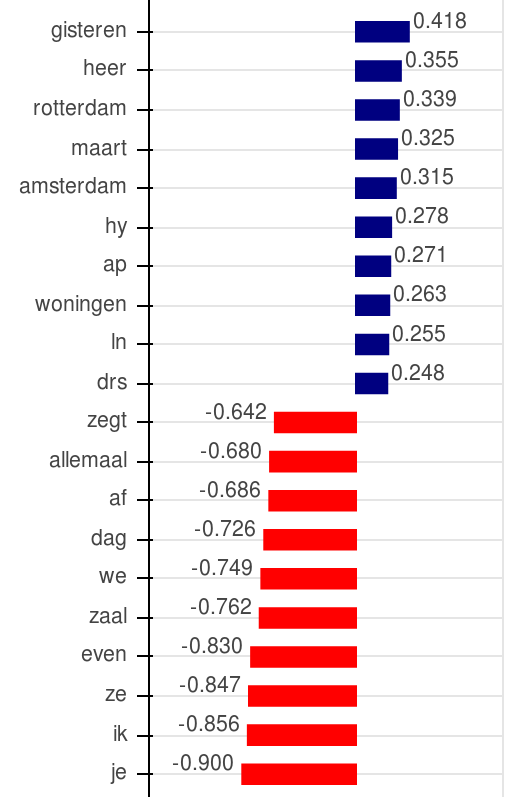}
\caption{Feature importance ranking plots comparing the top 10 features found indicative for \textit{News} (blue) in comparison with \textit{Feature} (red). The left graph shows the result of the pipeline SVC LIN 3 (GS (TF-IDF)) which includes stop words while the right graph belongs to the pipeline SVC LIN 3 (GS (TF-IDF) (SWR)) where stop-words are removed. The latter gives  more meaningful explanations thereby providing a more solid ground for analysis.}
\label{fig-nieuws-vs-rep}
\end{figure}

By utilizing the feature importance ranking plots for the SVC pipelines, we can get further insights on what the models deem relevant in general. When we look at the global explanations for \textit{Feature} given by the pipeline SVC LIN 3 (GS (TF-IDF)), we immediately see some interesting patterns. To start with, the recurrence of personal pronouns is interesting and relevant (see Fig.~\ref{fig-nieuws-vs-rep}). The emphasis on subjective experience in \textit{Feature} is in concordance with the understanding of the genre from a journalism history perspective. Therefore, the fact that the pipeline regards personal pronouns as a distinguished feature is considered to be encouraging. Moreover, the occurrence of personal pronoun {\lq}ik{\rq} (I) in \textit{Feature} is identified as indicative in comparison with \textit{Review}, \textit{Background}, \textit{Report} and \textit{News}, but not as indicative in comparison with \textit{Interview}, \textit{Column} and \textit{Op-ed}. This is a favorable pattern as \textit{Interview}, \textit{Column} and \textit{Op-ed} are also genres that focus on either the experience of the journalist or of the subject. The \textit{Column}, for example, focuses even more on the author of the article than the \textit{Feature} does. The features that are found to be important for \textit{Feature} in comparison with these genres are claimed to be certainly interesting as they reinforce the journalism historian{\rq}s understanding of the genre \textit{Feature} by pointing at the description of a particular event or moment through the usage of words such as {\lq}zaal{\rq} (hall; Fig. \ref{fig-nieuws-vs-rep}), {\lq}dag{\rq} (day; Fig. \ref{fig-rep-vs-col}) and {\lq}uur{\rq} (hour; Fig. \ref{fig-rep-vs-col}). 

We observe another relevant pattern with the occurrence of {\lq}gisteren{\rq} (yesterday) that can be seen in both of the feature importance ranking plots for \textit{News} (the graph on the right in Fig. \ref{fig-nieuws-vs-rep}) and for \textit{Report} (the graph on the right in Fig. \ref{fig-rep-vs-col}). According to journalism history literature, these genres both report on newsworthy events in which context the usage of {\lq}gisteren{\rq} becomes a meaningful feature. These findings help the journalism historian bridge the gap between their abstract understanding of genres and the concrete features needed to automatically classify them. 

In line with our expectations, not all findings are favorable by the journalism historians. However, we note them in Section \ref{subsec-lessons} where we provide a detailed discussion around lessons learned by the help of transparency.

\subsubsection{Local interpretations}

To shed some light on the reasoning of the pipelines, we turn to local interpretations based on the individual articles to improve our understanding.

First, we look at two articles that are correctly classified by the three pipelines and compare their explanations in order to determine whether the genre classification is based on relevant features from a journalism history perspective.  

\begin{figure*}[t]
\centering
\frame{\includegraphics[scale=0.211]{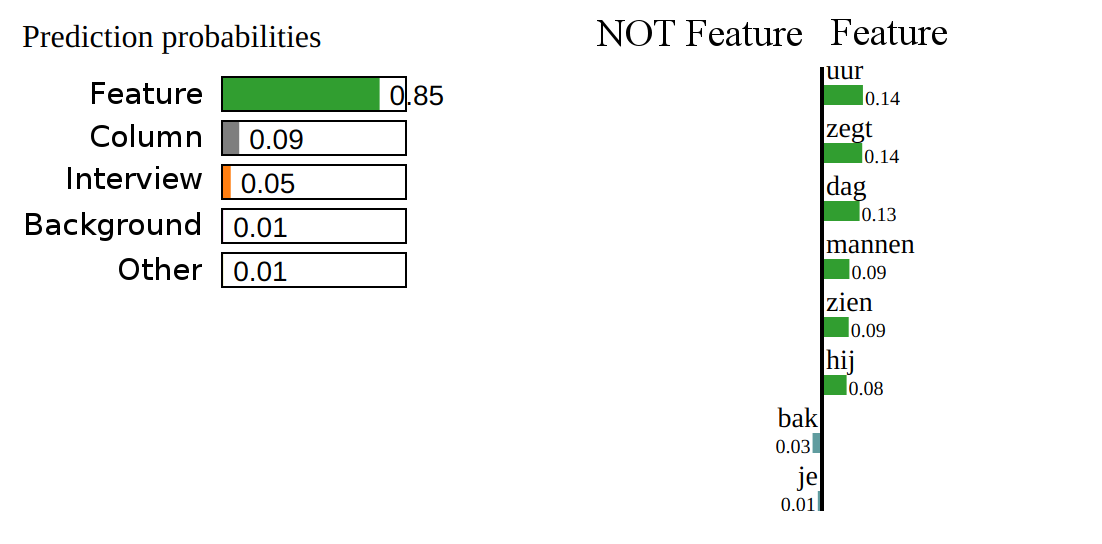}}
\frame{\includegraphics[scale=0.145]{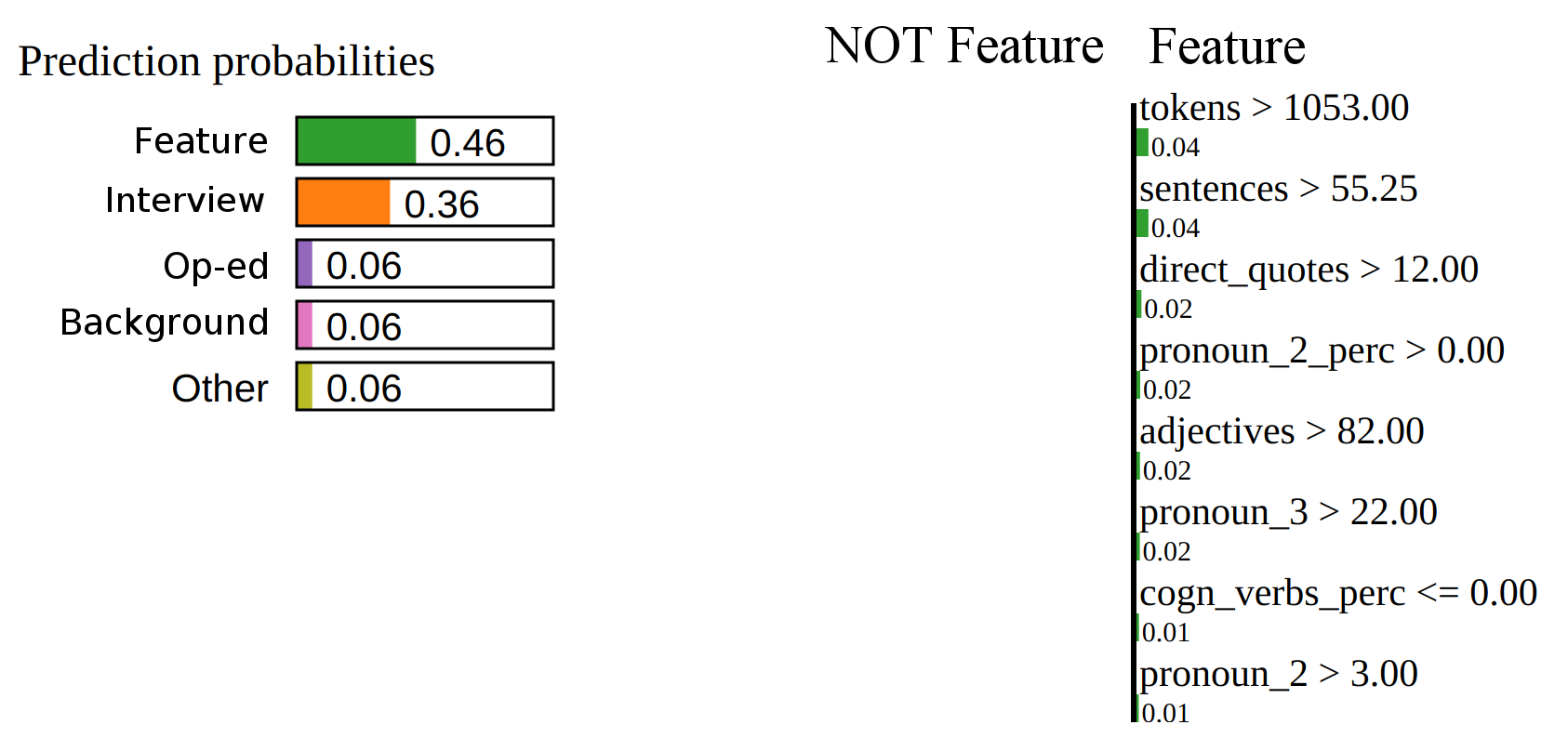}}
\caption{Local explanations for Article 239 given by the pipeline (left box) SVC LIN 3 (GS (TF-IDF) (SWR)) and (right box) RF 50 5 (BGS (FROG) (SCL)) [Images produced by LIME \cite{ribeiro2016should}]}
\label{fig-lime239}
\end{figure*}

\begin{itemize}
\item The local explanations given by two pipelines for Article 239, which has a ground truth label as \textit{Feature}, can be seen in Fig. \ref{fig-lime239}. For both explanations (within boxes in Fig. \ref{fig-lime239}), the probability for the top four predictions can be seen on the left and the predictive features for the class that the pipeline has predicted, in this case \textit{Feature}, are listed on the right. The SVC pipeline (left box) shows a focus on the perception of a certain moment, highlighting {\lq}uur{\rq} (hour), {\lq}dag{\rq} (day), and {\lq}zien{\rq} (seeing). Its textual explanations does not show topic words. The feature explanation of RF 50 5 (BGS (FROG) (SCL)) shows some features that can indeed be related to the characteristics of \textit{Feature}. From a journalism history perspective, the large number of adjectives (Fig. \ref{fig-lime239}, right), for example, indicates the focus on the atmosphere and subjective description. The large number of sentences and the number of tokens are also well-known indicators for \textit{Feature}, which tends to have comparatively longer articles together with \textit{Background} and \textit{Interview} in relation to the articles belonging to the other genres. 

\item The local explanations given by various pipelines for Article 189 has a ground truth label as \textit{Report}. Both SVC pipelines show that the article has been classified as \textit{Report} based on its topic. A large part of the top features having the most importance is related to the act of playing chess: {\lq}zwart{\rq} (black), {\lq}verdediging{\rq} (defense), {\lq}partij{\rq} (game), {\lq}ronde{\rq} (round), {\lq}spel{\rq} (game) and {\lq}remise{\rq} (draw). The explanations given by the pipeline RF 50 5 (BGS (FROG) (SCL)) are more informative and relevant to our understanding of \textit{Report} from a journalism history perspective. We observe that the lack of first and second personal pronouns are strong indicators for the prediction. Since \textit{Report} is an impersonal, factual, chronological description of an event, these characteristics are perceived to be robust features for genre classification. 
\end{itemize}

Next, we look at an article that is misclassified by the SVC pipelines, which exhibit the highest accuracy scores in the environment. 

\begin{itemize}
\item Article 175 has a ground truth label as \textit{Interview} and is classified correctly by the pipeline RF 50 5 (BGS (FROG) (SCL)). The local explanations retrieved from LIME for this pipeline show high relevancy with this genre. They include direct quotes, remaining quotation marks, the amount of first person pronouns and (relative) amount of third person pronouns, which are all distinguished characteristics of \textit{Interview}.

On the other hand, the SVC pipelines misclassified Article 175 and they both predicted it as \textit{Op-ed}. When we look at the local explanations of SVC LIN 3 (GS (TF-IDF)), we observe topic words, such as {\lq}god{\rq}, {\lq}samenleving{\rq} (society), and {\lq}communisten{\rq} (communists), which indicate that the pipeline associates political-philosophical topics with the genre \textit{Op-ed} and therefore \textit{Interview} on the same topic is misclassified. From a journalism history perspective, identifying genres through topic is not appealing as topic is not understood as an intrinsic part of genres, while textual form is.

\end{itemize}

Lastly, we look at an unlabeled article (from the data set UD) on which the three pipelines disagree. Looking at the raw text of the article, we classified Article 60 as \textit{Review}. Both SVC pipelines provide a correct prediction whereas the RF pipeline predicts that Article 60 is a \textit{Background} article. After carefully examining the local explanations of RF 50 5 (BGS (FROG) (SCL)), the prediction seems to be mainly based on the amount of tokens and sentences combined with lack of personal pronouns and lack of exclamation marks. These are indeed relevant characteristics of \textit{Background}. This observation signals that the RF pipeline did not have enough examples to learn the distinction between \textit{Background} and \textit{Review}. Concerning both of the SVC pipelines, the local explanations demonstrate a tendency towards topic classification. From a journalism history perspective, this tendency can be justified due to the fact that \textit{Review} is very closely connected to a specific topic and it can be regarded as an opinion article about cultural products.

\subsection{Domain hypotheses testing}
Based on their exploration of the accuracy scores, the confusion matrices, together with the global and local explanations, the domain scientists can make informed decisions regarding which pipeline is better suited for hypotheses testing. The pipeline that can distinguish genre characteristics rather than topic, and that is as accurate as possible would be the best candidate. 

\begin{figure}[t]
\includegraphics[scale=0.25]{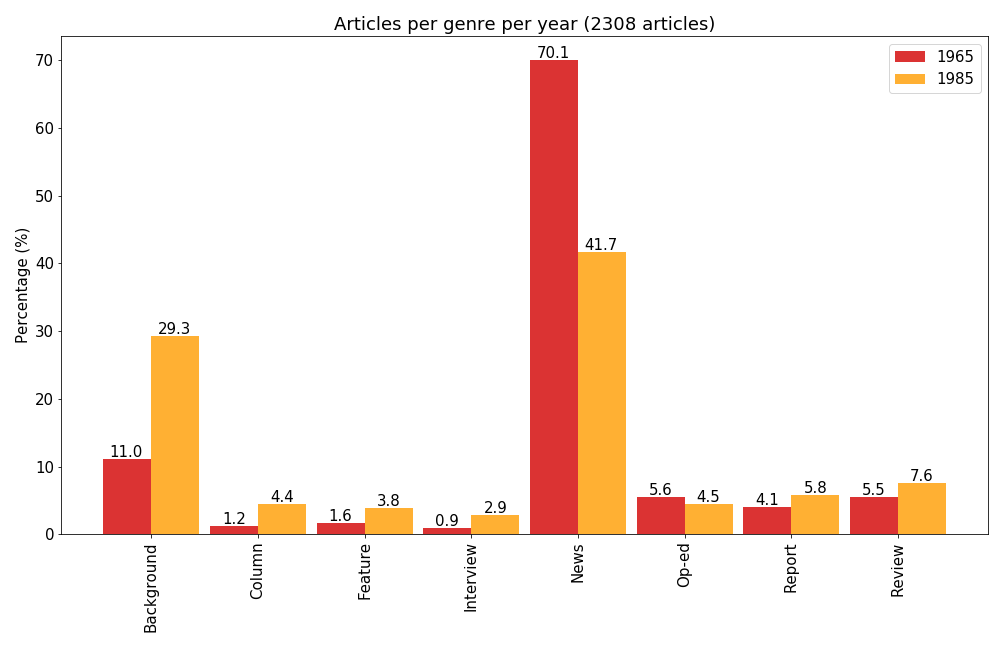}
\caption{Genre frequency distribution in the Dutch data of Harbers \cite{harbers2014}. In comparison with articles in the newspaper NRC from the year 1965, the NRC articles of the year 1985 contained more of the genres \textit{Background}, \textit{Column}, \textit{Feature}, \textit{Interview}, \textit{Report} and \textit{Review}, and less of the genres \textit{News} and \textit{Op-ed}.}
\label{fig-1965-vs-1985}
\end{figure}

\begin{figure}[t]
\includegraphics[scale=0.25]{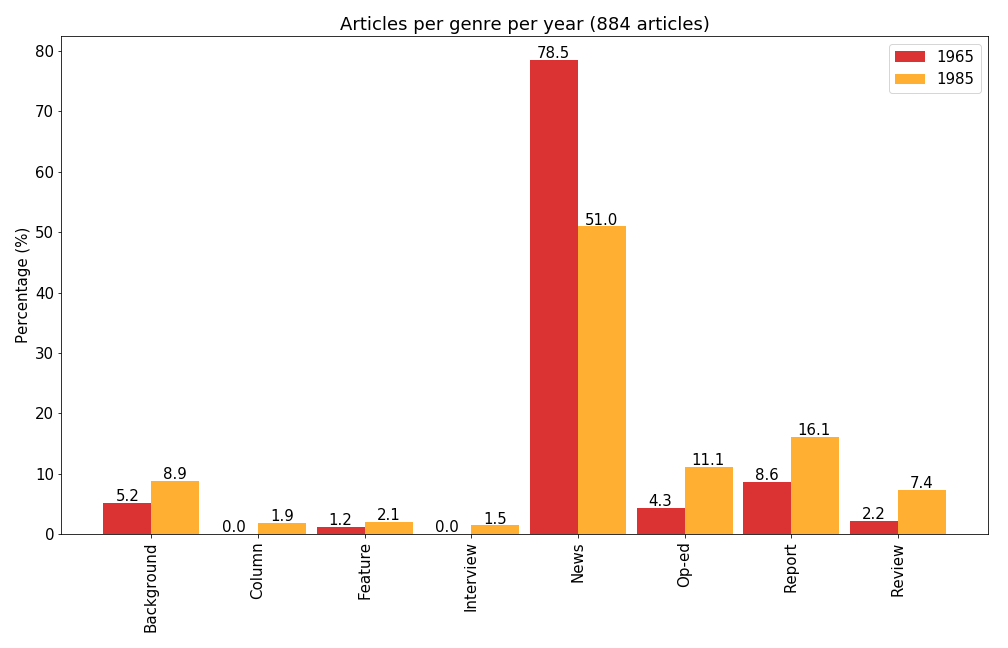}
\caption{Genre frequency distribution in the output of the pipeline RF 50 5 (BGS (FROG) (SCL)) for unseen articles of the newspaper NRC. The heights of the bars were recomputed to accommodate for the errors of the machine learning model. There are striking differences with the distributions of \textit{Background}, \textit{Op-ed}  and \textit{Report} in Fig. \ref{fig-1965-vs-1985} but the two hypotheses given in Section \ref{sec-hypotheses} are satisfied.}
\label{fig-recomputed}
\end{figure}

After further investigation on also the other pipelines that are among the remaining six (see Section \ref{subsec-choosingpipeline} for the details), the pipeline RF 5 50 (BGS (FROG) (SCL)) was chosen to be the best candidate to perform the hypotheses testing since the interpretations based on the features are more convincing, closer to the journalism historians{\rq} abstract understanding of the genres and, most importantly, not related to the topic of the articles. Fig. \ref{fig-recomputed} shows the hypothesis graph based on RF 5 50 (BGS (FROG) (SCL)). We observe that the results are compliant with both hypotheses: the amount of \textit{Feature} and \textit{Report} both increase in 1985.
However, if we compare Fig. \ref{fig-recomputed} with the genre distribution graph in Fig. \ref{fig-1965-vs-1985} based on the manually labeled date by Harbers \cite{harbers2014}, we see that the RF 5 50 (BGS (FROG) (SCL)) is overestimating the relative amount of \textit{Reports} and \textit{Op-eds} and underestimating the amount of all other genres, especially \textit{Background}. This comparison shows us that even though RF 5 50 (BGS (FROG) (SCL)) is the best pipeline available in terms of feature explanation (Fig. \ref{fig-lime239}) and meaningful genre categorization (Fig. \ref{fig-cm}), future work is needed to investigate pipelines that are sufficiently reliable and accurate.

\begin{figure}[t]
\includegraphics[scale=0.23]{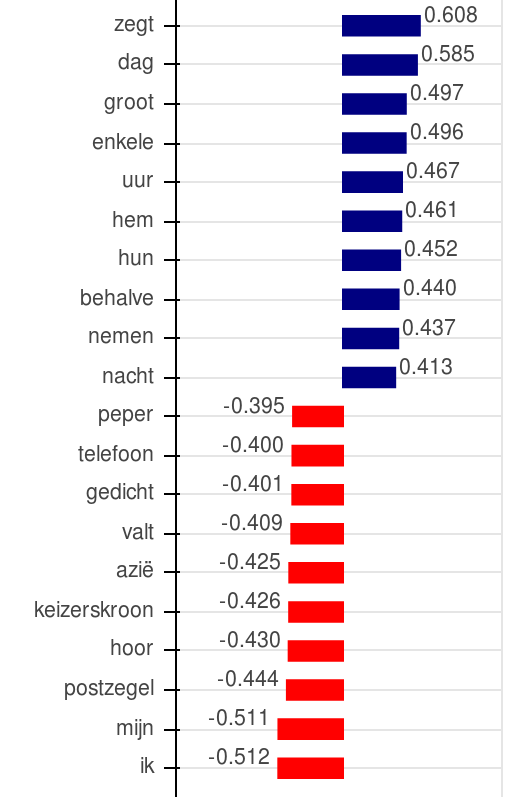}
\includegraphics[scale=0.23]{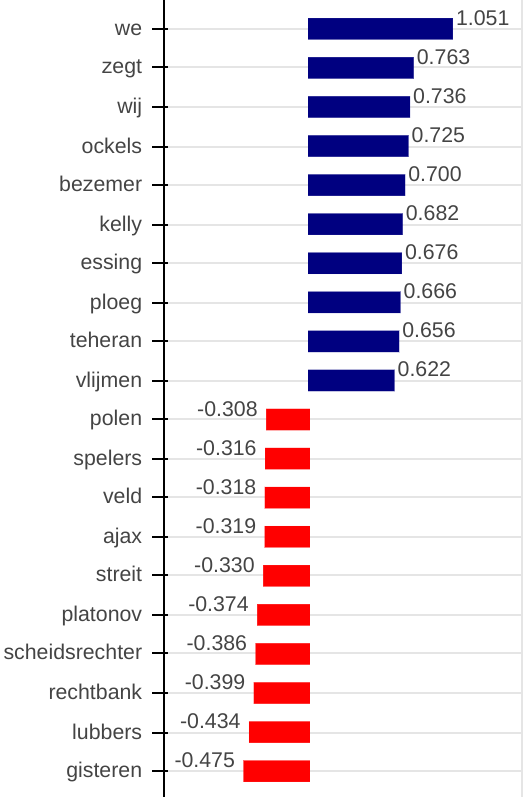}
\caption{Feature importance ranking plots given by the pipeline SVC LIN 3 (GS (TF-IDF) (SWR)) comparing the top 10 features found indicative for (left) \textit{Feature} (blue) in comparison with \textit{Column} (red); (right) \textit{Interview} (blue) in comparison with \textit{Report} (red).}
\label{fig-rep-vs-col}
\end{figure}

\section{Discussion}
\label{sec-discussion}
\subsection{Building a transparent environment}

From a computer science perspective, one of the notable challenges of building a transparent environment is the depth of domain relevant knowledge exposure. In other words, how much of the algorithmic details should be shown and to what extent they can be absorbed by the journalism historian is a relevant planning to make during the design process. One way to facilitate the navigation through the details of the data science domain is to provide guiding logic in creation of the pipeline. For example, although it may be trivial to the data scientist not having to select features from a TF-IDF representation, this needs to be explicitly presented to the journalism historian when constructing the pipeline. Thus, it is crucial to address domain related knowledge gaps especially when non-computer scientists are handed the liberty to create their own pipelines for comparative analysis. 

During the integration of pre-processing options (e.g. selection of text representation, applying stop-word removal to the text, etc.), we realized that it is important to note the difference between the application of stop-word removal before and after parsing the text with the natural language processing tools as the removal of stop-words may influence the parse-tree. Obviously, this case would be of concern under the circumstances where the representation of the text relies on the natural language processing tools that provide the parse-tree. Similarly as above, the environment needs to communicate this to the journalism historian. We leave the impact analysis of this observation for future work.

The pre-processing of the data, evaluation of classification algorithms and their hyper-parameters can get quite expensive in terms of computational requirements and processing times depending on the number of options considered in the environment. In Step 1 of the workflow, we considered two different natural language processing (NLP) tools available for Dutch language, which are SpaCy\footnote{\url{https://spacy.io/}} and Frog. Although we believe that measuring the impact of each step in an ML pipeline is a prerequisite to claim transparency, we leave the impact of using different NLP tools out of scope of this paper. With regards to Steps 2 and 3 in the workflow, it is important to acknowledge the potential impact of various algorithms, parameters and scorers. For example, we looked at 4 scorers during the grid search in Step 2. However, we note that this is not an exhaustive search, but rather a jump-start for the more advantageous numerical performance, which is commonly perceived as the major determinant in selection of an appropriate machine learning pipeline for the journalism historian's task.

\subsection{Choosing the best candidate pipeline: a cyclic process}
\label{subsec-choosingpipeline}

As noted by the journalism historians, choosing the best candidate pipeline to be employed for hypothesis testing is not straightforward. Although the transparency provided through the global and local explanations lead to some favorable pipelines, the domain scientist still carries doubts about the accuracy scores being low. The explanations reveal the underlying reasoning that the pipelines make, and if the journalism historians do not detect alignment between the journalistic genre definitions and what the ML pipeline has learned, then they will prefer pipelines with lower accuracies.

Having such hesitation ignited a cyclic approach between Steps 4 to 6 in this study. While the determination of the best performing pipeline was at first based on mainly the accuracy score and the confusion matrices, the knowledge gained through navigating in the environment caused the journalism historian to look beyond the accuracy score. To a non-computer scientist, the environment opened new doors to reconsider the impact of the chosen algorithm, the training data set, and the pre-processing on the performance of the pipeline. 
Under this light, we also looked at other pipelines. Within the remaining six, the pipeline RF 1000 5 (GS (FROG) (SCL)) is worth investigating further as it uses the same representation type as the pipeline RF 50 5 (BGS (FROG) (SCL)) and has an appealing accuracy score. Although the explanations of RF 50 5 (BGS (FROG) (SCL)) are sometimes generic, the explanations of RF 1000 5 (GS (FROG) (SCL)) in most cases are more specific. For example, the latter has classified Article 239 correctly as \textit{Feature}. The feature explanation graphs show that amount of tokens, third person pronouns, adjectives, sentences, pronouns, remaining quotes, intensifiers and subjectivity are the most important ones. This is a valid explanation as it indicates that \textit{Feature} is distinguished to be a long, subjective article, with relatively many adjectives, third person pronouns, first person pronouns and intensifiers. For journalism historians, these characteristics are parallel to their understanding of \textit{Feature}. The features that they consider crucial such as subjectivity, adjectives, and intensifiers, which are very typical for the subjective description in \textit{Feature}, are as well recognized by the pipeline.

Having had further insight regarding the potential impact of the data set of the pipeline on a local level, it was only tempting to explore this on the global level. For this purpose, we looked at the feature importance rankings given by the pipeline SVC LIN 2 (BGS (TF-IDF) (SWR)). These plots were compared to the global explanations given by the pipeline LIN 3 (GS (TF-IDF)). Subsequently, the particular investigation on the distinction between \textit{Feature} and \textit{Column} was recorded to be entirely different in these two pipelines. The pipeline SVC LIN 2 (BGS (TF-IDF) (SWR)) was found to be showing barely topic-related features, and in fact, it identified \textit{Column} to be about the columnist through the use of first person pronouns. 

Going back to the local level for the pipeline SVC LIN 2 (BGS (TF-IDF) (SWR)) was slightly disappointing as a visible tendency for topic classification was found. Article 189, for example, was correctly classified as \textit{Report} by the pipeline. The most important features for the prediction are {\lq}partij{\rq} (team), {\lq}hij{\rq} (he), {\lq}gisteren{\rq} (yesterday), {\lq}zetten{\rq} (to move or the moves), {\lq}tilburg{\rq} (Tilburg (city name)), {\lq}afgebroken{\rq} (cancelled), which are partly related to the topic chess. Although the focus on topic is considerably less clear in comparison with the local explanations given by the pipeline SVC LIN 3 (GS (TD-IDF)), the occurrence of {\lq}partij{\rq} (team) and {\lq}zetten{\rq} (moves) still signify that predictions are based on topic.

The cyclic approach iteratively led the journalism historians to think that using a pipeline based on the BGS data set and linguistic features is preferable. As a consequence of the transparency support provided by the environment, journalism historians ended up at a totally different point of view where they were freed from \textit{}the bias of high accuracy scores and they in fact chose the pipeline with the lowest accuracy score to continue their hypotheses testing with. This decision was made based on increased trust and understanding.

\subsection{Lessons learned with the aid of transparency}
\label{subsec-lessons}

In journalism history literature, some genre-topic combinations are more frequent than others. For example, \textit{Review} is a genre that always has cultural products as a topic while \textit{Report} is much more often about sport events or court hearings (in the time frame under consideration in this paper). The findings of this study reveal that the pipelines with high accuracy are likely to perform topic classification. This can be visualized in the confusion matrices, especially in those which have darker cells at the diagonal for genres such as \textit{Review} and \textit{Report}.

As postulated in computer science literature, global interpretability implies general patterns between genres. However, our findings do not comply with this statement. For example, the feature importance ranking of \textit{Feature} versus \textit{Column} in Fig. \ref{fig-rep-vs-col}, shows that some of the most important features for \textit{Column} are {\lq}peper{\rq} (pepper), {\lq}telefoon{\rq} (telephone), {\lq}gedicht{\rq} (poem), and {\lq}postzegel{\rq} (stamp). These words together convey a certain focus on homely and personal event. Although columns often tend to be about these topics, this is not a valid generalization on the genre. 

Although on a global level the tendency for topic classification is certainly less visible, on a local level some articles still show how topic plays into the (albeit right) classification. It can be deduced that the data set of the pipeline certainly has an effect, but even a balanced (across genres) data set does not completely alleviate the tendency of the pipelines that use TF-IDF representation to perform classification based on topic rather than genre-related characteristics. On the other hand, the global explanations given by the pipelines that use FROG representation show favorable compliance with the genre-related characteristics, which are rooted in the journalism history literature. In our study, this is most noted on \textit{Feature} versus all the other genres, where the global explanations are consistently related to the most important characteristics of the \textit{Feature} as a genre.

\section{Concluding remarks and future work}
\label{sec-conclusion}

In this paper, we demonstrated the impact of a transparency-led environment that facilitates machine learning pipeline selection for non-computer scientists using a case study on automatic genre classification in journalism history. We noted that transparency-based exploration indeed made a change in the journalism historians{\rq}{} decision making criteria when choosing a machine learning pipeline for hypothesis testing.

Furthermore, we detected the potential to gain valuable insights into how the genre definitions from the journalism history literature can be connected to the relevant features mined from the data sets (see also Broersma \cite{broersma2018}). Although claiming to identify genre-related characteristics that are not found in the literature yet is a far-stretching effort at the current maturity of this study, we perceive this as a remarkable advantage in the future of transparency-driven eSciences.

As part of future work, we plan to investigate the impact of diversifying the design decisions within the tools used in each of the steps of the ML pipelines. This includes, for example, trying improved version of LIME that gives higher precision on unseen instances \cite{ribeiro2018anchors}, integrating different NLP tools while also applying different NLP configurations, etc. Furthermore, we are interested in the agreement between the collective local explanations (retrieved from the entire data set of the pipeline) and global explanations and whether this will improve the scientific understanding of the domain scientists. Finally, in this paper, we have primarily focused on potential bias and errors in the machine learning models in order to answer our domain-focused research questions. In the future, we also want to investigate the impact of biases that emerge in the rest of the pipeline such as data selection, data annotation, use of different NLP tools, etc.
 
\section*{Acknowledgments}
The study described in this paper was executed as a part of the NEWSGAC project which is funded by the Netherlands eScience Center\footnote{esciencecenter.nl} and CLARIAH\footnote{clariah.nl}.

% references section
\bibliographystyle{plain}
\bibliography{bibliography}

\end{document}